\documentclass[final,5p,times]{elsarticle}
\usepackage{graphicx}
\usepackage{subcaption}
\usepackage{epsfig}
\usepackage{siunitx}
\usepackage{booktabs}
\usepackage[T1]{fontenc}
\usepackage{multirow} 
\usepackage{tabularx}
\usepackage{array}
\usepackage[namelimits]{amsmath} 
\usepackage{amssymb} 
\usepackage{amsmath}            
\usepackage{amsfonts}            
\usepackage{mathrsfs} 
\usepackage{float}
\usepackage{cases}
\usepackage{wrapfig}
\usepackage{hyphenat}
\usepackage{multicol}
\usepackage{caption}
\usepackage[labelfont=bf, textfont=normalfont, justification=raggedright, singlelinecheck=false]{caption}
\usepackage{graphicx}
\usepackage{subcaption}
\usepackage{tabularx}

\usepackage{amssymb}

\journal{arxiv}
\date{}

\begin{document}
\begin{frontmatter}



\title{A Highlight Removal Method for Capsule Endoscopy Images \tnoteref{13}}


\tnotetext[13]{This work was supported in part by the National Natural Science Foundation of China (No. 62172190), National Key Research and Development Program(No. 2023YFC3805901), the "Double Creation" Plan of Jiangsu Province (Certificate: JSSCRC2021532) and the "Taihu Talent-Innovative Leading Talent" Plan of Wuxi City(Certificate Date: 202110). }
\author[label1]{Shaojie Zhang}
\ead{7213107006@stu.jiangnan.edu.cn}
\author[label1,label2]{Yinghui Wang\corref{cor1}}
\ead{wangyh@jiangnan.edu.cn}
\author[label1]{Peixuan Liu}
\ead{362342276@qq.com}
\author[label1]{Wei Li}
\ead{cs_weili@jiangnan.edu.cn}
\author[label1]{Jinlong Yang}
\ead{yjlgedeng@163.com}
\author[label1]{Tao Yan}
\ead{yantao.ustc@gmail.com}
\author[label1]{Yukai Wang}
\ead{ericwangyk22@163.com}
\author[label3]{Liangyi Huang}
\ead{lhuan139@asu.edu}
\author[label4]{Mingfeng Wang}
\ead{mingfeng.wang@brunel.ac.uk}
\author[label5]{Ibragim R. Atadjanov}
\ead{ibragim.atadjanov@gmail.com}
\cortext[cor1]{Corresponding author}
\affiliation[label1]{organization={ School of Artificial Intelligence and Computer Science, Jiangnan University},
            addressline={1800 Li Lake Avenue},
            city={wuxi},
            postcode={214122},
            state={Jiangsu},
            country={PR China}}
\affiliation[label2]{organization={ Engineering Research Center of Intelligent Technology for Healthcare, Ministry of Education},
            addressline={1800 Li Lake Avenue},
            city={wuxi},
            postcode={214122},
            state={Jiangsu},
            country={PR China}}
 \affiliation[label3]{organization={School of Computing and Augmented Intelligence, Arizona State University},
            addressline={1151 S Forest Ave},
            city={Tempe},
            postcode={8528},
            state={AZ},
            country={U.S}}
\affiliation[label4]{organization={Department of Mechanical and Aerospace Engineering, Brunel University},
            addressline={Kingston Lane},
            city={London},
            postcode={UB8 3PH},
            state={Middlesex},
            country={U.K}}   
\affiliation[label5]{organization={Tashkent University of Information Technologies named after al-Khwarizmi},
			addressline={ 108 Amir Temur Avenue},
		    city={Tashkent},
			postcode={100084},
			state={},
			country={Uzbekistan}}

\begin{abstract}
The images captured by Wireless Capsule Endoscopy (WCE) always exhibit specular reflections, and removing highlights while preserving the color and texture in the region remains a challenge. 
To address this issue, this paper proposes a highlight removal method for capsule endoscopy images. 
Firstly, the confidence and feature terms of the highlight region's edges are computed, where confidence is obtained by the ratio of known pixels in the RGB space's R channel to the B channel within a window centered on the highlight region's edge pixel, and feature terms are acquired by multiplying the gradient vector of the highlight region's edge pixel with the iso-intensity line. 
Subsequently, the confidence and feature terms are assigned different weights and summed to obtain the priority of all highlight region's edge pixels, and the pixel with the highest priority is identified. 
Then, the variance of the highlight region's edge pixels is used to adjust the size of the sample block window, and the best-matching block is searched in the known region based on the RGB color similarity and distance between the sample block and the window centered on the pixel with the highest priority. 
Finally, the pixels in the best-matching block are copied to the highest priority highlight removal region to achieve the goal of removing the highlight region. Experimental results demonstrate that the proposed method effectively removes highlights from WCE images, with a lower coefficient of variation in the highlight removal region compared to the Crinimisi algorithm and DeepGin method. Additionally, the color and texture in the highlight removal region are similar to those in the surrounding areas, and the texture is continuous.\end{abstract}




\end{frontmatter}


\section{Introduction}
The use of Wireless Capsule Endoscopy (WCE) can effectively alleviate patient discomfort while comprehensively obtaining relevant information about the patient's digestive tract \cite{ref1}. However, due to the smooth surface of the gastrointestinal tract and the proximity of the WCE light source to the gastrointestinal surface, WCE images often exhibit intense specular reflections, resulting in highlight regions where gastrointestinal information is lost. Nevertheless, the highlight regions in gastrointestinal images may contain crucial information related to the gastrointestinal tract, such as color, texture, and especially lesions \cite{ref2}. Without highlight removal, it can interfere with doctors' observation and judgment of lesions, making it challenging to identify hidden abnormalities and thereby reducing diagnostic accuracy \cite{ref3}.
Compared to extracavitary images, WCE images exhibit a unique phenomenon of intense specular reflections and a relatively large proportion of highlight areas due to the alignment of the light source illumination direction with the camera's imaging direction. This is further exacerbated by the wet and smooth surface of the gastrointestinal tract, where the light source and camera are in close proximity to the inner wall of the gastrointestinal tract. Since Sharfer introduced the bidirectional reflectance distribution function (BRDF) model \cite{ref4}, many model-based methods have been developed for highlight removal. While these methods are effective on natural images, their applicability in endoscopic scene images is often limited, as certain conditions satisfied by natural images cannot be met in endoscopic images.
In contrast, highlight removal methods based on image restoration \cite{ref9}-\cite{ref17} have been explored in the medical field, showing good adaptability when dealing with small-area highlights. For larger highlight regions, texture-based image restoration techniques have been employed, wherein a pixel at the damaged image edge is selected as the core, and existing pixel blocks in the image are matched to fill the region with missing information, achieving image recovery. The Criminisi algorithm \cite{ref13} is a representative method in this category. This algorithm considers local contextual information, allowing for precise restoration based on complex texture features. It efficiently reconstructs various sizes of damaged areas, excelling in the treatment of extensive information loss and displaying strong adaptability to noisy images. However, improper selection of restoration sequence and matching strategy in this algorithm may result in issues such as incorrect texture continuation and texture disjunction.
In recent years, some scholars have proposed deep learning-based methods \cite{ref18}-\cite{ref21} for removing specular reflections and highlights in images. However, these methods often require high-quality training datasets to ensure robustness and universality, and the high difficulty in obtaining annotated data for the human body's internal cavities limits the generalizability of deep learning-based approaches.
Overall, in the context of the human body's internal environment, there is a stronger correlation between adjacent pixels in images \cite{ref23}-\cite{ref25}. Additionally, due to the presence of hemoglobin in the human body, the R-channel value in WCE images is significantly greater than the B-channel value in the RGB space. When WCE images are subjected to intense highlight or insufficient lighting, the R-channel and B-channel values become closer \cite{ref26}. To address this, we have enhanced the Criminisi algorithm to leverage its restoration advantages and enable effective highlight removal in WCE images. To better represent the known information within the restoration window, we use the ratio of R-channel to B-channel as confidence, introducing more known information into the Criminisi algorithm \cite{ref13}. Subsequently, to emphasize the role of confidence and feature terms, we calculate the priority at the highlight edge by adding the confidence and feature term weights, ensuring correct restoration order and linear structure propagation. Finally, by adjusting the sample block window size based on the local variance of the highlight region's edge and incorporating pixel distance factors in the calculation of the best-matching block, we obtain more accurate matches. This approach effectively removes specular reflections and highlights in WCE images, ensuring similarity in color and continuous texture in the highlight removal region and its surroundings.
This paper's main contributions can be summarized as follows:

(1) Developed a pixel priority calculation method supporting linear structure propagation. Utilizing the R-channel to B-channel ratio in WCE images' RGB color space, we refined confidence in pixel priority calculation, ensuring accurate assessment of each pixel's repair potential and maintaining precision in highlight restoration sequence within the human body's internal environment. 

(2) Improved the calculation of confidence and feature terms in pixel priority computation. Specifically, a weighted sum strategy was employed to avoid issues where high confidence in a pixel, coupled with zero feature terms, could result in its neglect during prioritized restoration.

(3) Introduced a sample block-based adaptive matching met\\hod. Dynamically adjusts the sample block window size using variance and optimizes the selection of the best-matching block within the window by introducing distance factors, enhancing the accuracy of the search for the best match.

\begin{figure*}[t]
    \centering
    \includegraphics[height=3in]{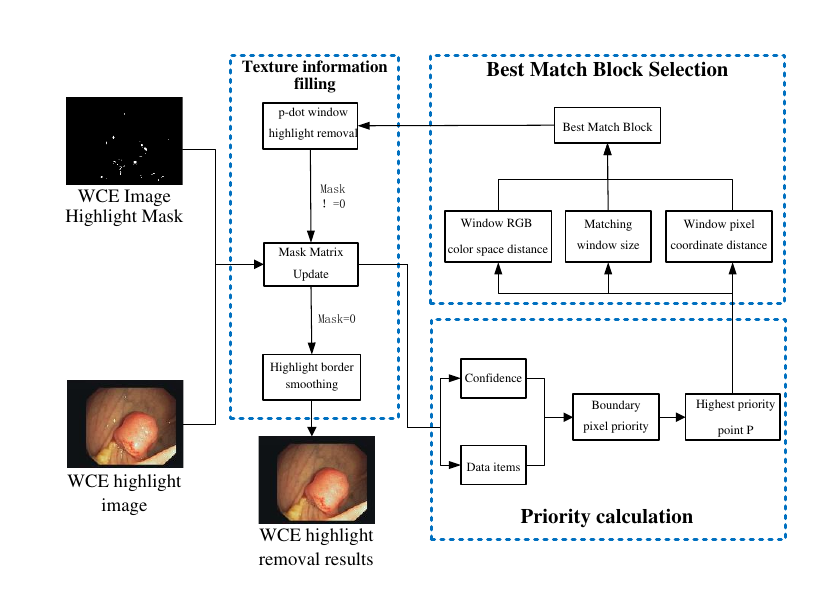}
\caption{Methodological Framework.}
    \label{Fig1}
\end{figure*}

\section{RELATED WORK}
Currently, highlight removal methods mainly include bidirectional reflectance distribution function (BRDF) models \cite{ref4}-\cite{ref8}, image restoration techniques \cite{ref9}-\cite{ref17}, and deep learning-based approaches \cite{ref18}-\cite{ref21}.

The bidirectional reflectance distribution function (BRDF) model is commonly used for image analysis and processing, aiming to separate reflection and illumination components. Shaf\\er \cite{ref4} introduced a BRDF model that decomposes highlight images into diffuse and specular reflection components. Subsequent variations include Tan \cite{ref5} Specular Free (SF) image, Shen \cite{ref6} improved SF image, and Yang \cite{ref7} approach using bilateral filtering. While effective, these methods often introduce chromatic changes or fail to completely remove specular reflections. Guo \cite{ref8} proposed a sparse low-rank reflection model, optimizing both diffuse reflection and specular highlight images. However, it may result in excessively dark pixels in specular highlight regions. Despite the effectiveness of the BRDF model, its application to WCE images reveals that the intense highlights in these images predominantly consist of specular reflection components. Consequently, after removing these components, the highlight removal region appears almost black.

Image restoration is the process of reconstructing missing or degraded regions in an image. Given that the highlight regions in WCE images are primarily composed of specular reflection components, which qualify as degraded areas, and considering the strong pixel correlation in image neighborhoods, image restoration becomes a crucial method for highlight removal.One category of image restoration methods involves the use of variational partial differential equations (PDEs). Bertalm\\io \cite{ref9} proposed the BSCB model, employing PDEs to propagate known region information along iso-intensity directions to restore the image. However, this method is unsuitable for repairing large-scale damage or capsule endoscopy images with complex textures. Shen \cite{ref10} introduced the Total Variation (TV) image restoration model, which utilizes gradient descent flow to smooth images while preserving edge characteristics. Nevertheless, this method struggles to effectively restore missing regions in capsule endoscopy images with curve-like structures. Tsai \cite{ref11} proposed a restoration model based on Mumford-Shah, but its inability to achieve smooth curves hinders adherence to the image connectivity principle.While PDE-based image restoration methods can generate reasonable texture samples for repairing small damaged areas, they often result in prolonged repair times and unclear images when dealing with larger missing regions \cite{ref12}. Given that highlight regions in WCE images typically involve substantial information loss over large areas, with complex textures forming curve-like structures, these methods fail to produce satisfactory restoration results.Another category of image restoration methods is texture-based, exemplified by Criminisi's algorithm \cite{ref13}. This algorithm uses sample blocks centered around edge pixels in the region to be repaired, searching for the best-matching block in intact areas according to a precomputed priority order, and directly copying it for restoration. However, inaccuracies in the priority order or best-matching block calculation during the restoration process can lead to discontinuities in texture and color differences. Various scholars have proposed improvements, such as Zhang \cite{ref14}, who introduced curvature and gradient information to enhance restoration accuracy. Nevertheless, this method results in unclear image details. Na \cite{ref15} used Markov random fields \cite{ref16} as the matching criteria for the Criminisi algorithm to improve texture details, but it is slow and unsuitable for complex image restoration. Barnes \cite{ref17} introduced the PatchMatch method, which uses a fast nearest-neighbor algorithm to search for the most similar matching block, reducing memory consumption and computation costs during the search process. However, this method has poor convergence and long computation times.In summary, to adapt image restoration for highlight removal in the intracavitary environment of the human body, improvements are needed in the restoration sequence and matching strategy to avoid issues such as continuous texture errors and discontinuity.

Deep learning-based highlight removal methods train models to automatically adaptively eliminate highlights. Unlike approaches based on bidirectional reflectance distribution function (BRDF) models and image restoration, learning-based methods require appropriate training sets and time-consuming network training. Zeng \cite{ref18} employed the pixel with the highest confidence value in each iteration to progressively fill missing areas. While this method can restore fine texture details in high-resolution images, it requires repetitive iteration and significant computational resources. Xie \cite{ref19} introduced a learnable attention map module onto the U-Net structure \cite{ref20} and constructed learnable bidirectional attention maps using forward and backward attention maps. Although this method effectively repairs irregular missing regions, there are color differences in the restored region and its surroundings. Funke \cite{ref21} proposed a generative adversarial network for removing specular highlights in endoscopic images. The network is trained using patches extracted from endoscopic videos, some with specular highlights and others without. While it can remove small specular highlights, it exhibits boundary artifacts in the removed highlight regions. DeepGin \cite{ref22} designed a spatial pyramid extended ResNet block, utilizing distant features for image restoration, and employed multiscale self-attention and back-projection techniques to enhance restoration. However, it introduces texture artifacts in the restored regions. Although learning-based methods have made significant strides in computer vision, difficulties in obtaining paired images with and without highlights limit their potential applications. This challenge is even more pronounced when obtaining training sets for intracavitary environment images.

In summary, existing highlight removal methods are not suitable for the human intracavitary environment. Hence, this paper proposes a method tailored for intracavitary highlight removal.

\section{METHODOLOGY}

\subsection{Pixel Priority Calculation}
The Criminisi algorithm \cite{ref13} first performs priority calculation, assigning a temporary priority value to the blocks to be repaired along the damaged boundary. 
The priority values determine the order of repair to ensure the propagation of linear structures and connectivity of target boundaries in the image. 
Figure 2 represents the symbolic diagram of the Criminisi algorithm, where $I$ is the image to be repaired, $\phi$  is the undamaged region in the image, $\Omega$ is the damaged region in the image, $\delta \Omega$ is the boundary between the known and damaged regions, $p$ is the pixel with the currently highest priority value, $\Psi _{p}$ is the window centered at $p$, 
$\bigtriangledown I \frac{1}{p}$ is the direction of iso-intensity lines at point $p$, and  $n_{p}$ is the gradient normal vector orthogonal to the boundary. 
The priority calculation at pixel point p in the Criminisi algorithm is given by Equation (1).

\begin{equation}
	\label{eq1}
	P(p)= C(p)D(p)
\end{equation}
 
 Where $C(p)$ represents the confidence at pixel point $p$, indicating the proportion of known region pixels in the window $\Psi _{p}$. Its expression is given by Equation (2). A higher numerical value for $C(p)$ indicates a greater amount of known region information, warranting higher priority for repair. $D(p)$ denotes the feature term at pixel point $p$, representing the product of the gradient normal vector $n_{p}$ and the iso-intensity lines $\bigtriangledown I \frac{1}{p}$. Its expression is shown in Equation (3). A larger value for $D(p)$ signifies clear intersections between the known region and the damaged region, justifying higher priority for repair.
\begin{equation}
	\label{eq2}
	C(p)= \sum_{q\in  \Psi _{p}\bigcap\Phi }C(q)/\left | \Psi _{p} \right |
\end{equation}

\begin{equation}
	\label{eq3}
	D(p)= \left | \bigtriangledown I \frac{1}{p}\cdot n_{p}\right |/\alpha
\end{equation}
Where $\left |\Psi _{p} \right | $ represents the area of the window $\Psi _{p}$, $\alpha$ is the normalization factor, with a value of 255.

\begin{figure}[!hptb]
    \begin{center}
    \includegraphics[width=2.1in]{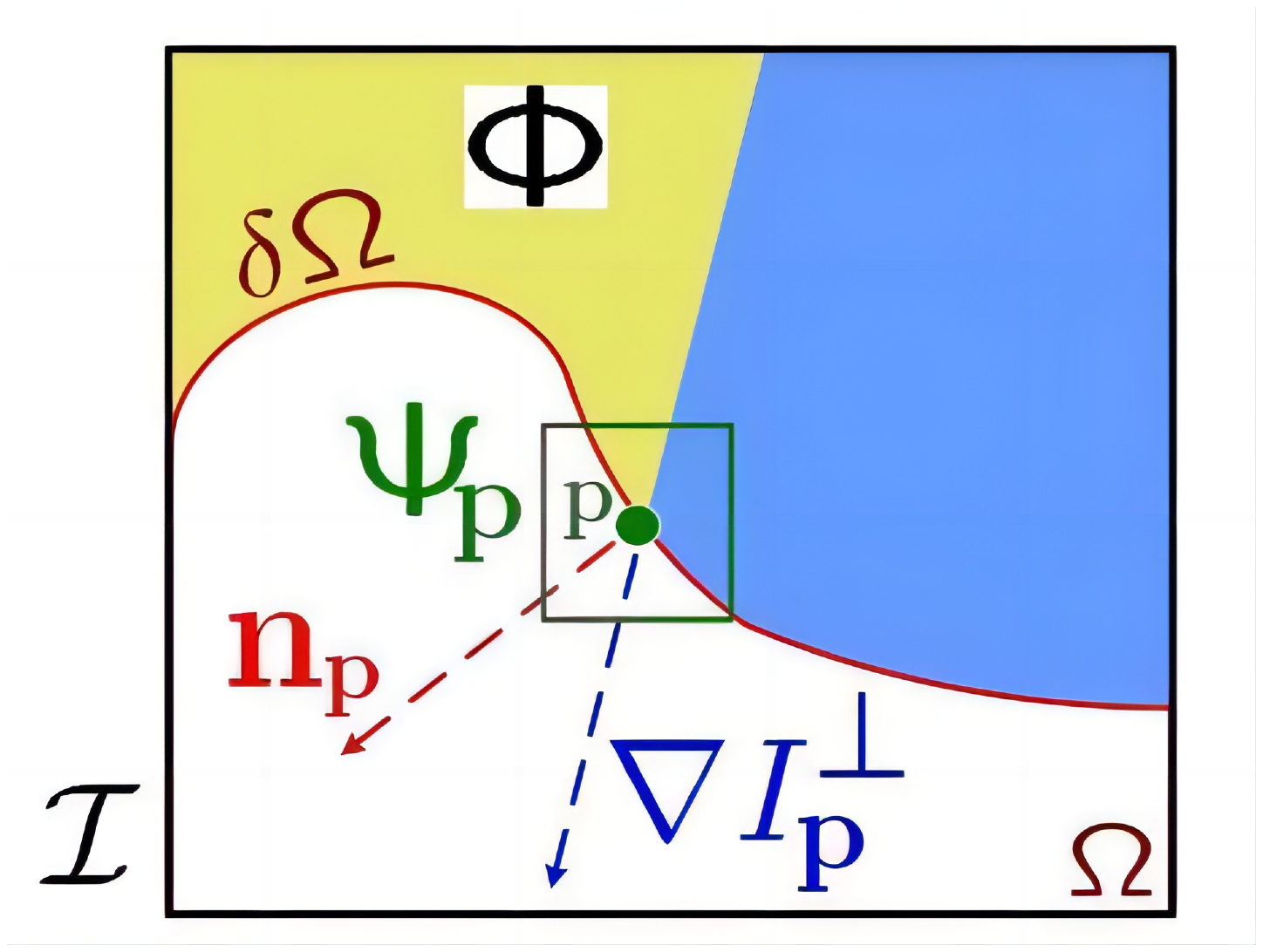}
    \end{center}
\caption{Symbolic Diagram of Criminisi Algorithm.} \label{Fig2}
\end{figure}

Due to the presence of hemoglobin, capsule endoscopy images appear red, with the R channel in the RGB space significantly greater than the B channel. The complex intracavitary environment and limited power of WCE result in both highlights and dark regions in the images. 
Through observation, it is noted that the highlight regions in WCE images, appearing as strong highlights close to white, have R channel and B channel values that are close \cite{ref23}. Conversely, dark regions in WCE images appear close to black, with R channel and B channel values also being close. 
To better reflect the confidence at the boundary of highlight regions in WCE images, 
this paper utilizes the ratio of the R channel to the B channel for known region pixels in the window $\Psi _{p}$ to represent the confidence at point $p$. A higher ratio indicates more valid information at point $p$. The improved confidence ${C}'(p)$ is expressed as Equation (4).

\begin{equation}
	\label{eq4}
	{C}'(p)= \sum_{q\in  \Psi _{p}\bigcap\Phi }RB(q)/\left | \Psi _{p} \right |
\end{equation}
Where $RB(q)$ represents the ratio of the R channel to the B channel for the pixel at point $q$, and its expression is given by Equation (5).

\begin{equation} 
	\label{eq5}
	RB(q)= R(q)/B(q)
\end{equation}	
Where $R(q)$ and $B(q)$ represent the R channel value and B channel value in the RGB color space for the pixel at point $q$.

When the iso-intensity lines  $\bigtriangledown I \frac{1}{p}$ are perpendicular to the gradient normal vector $n_{p}$, the feature term $D(p)$ becomes 0. This would prevent pixels with very high confidence from being prioritized for repair, leading to an incorrect priority order. To address this issue, this paper modifies the multiplication in Equation (1) to addition. The improved priority ${P}'(p)$ is expressed as Equation (6).
\begin{equation} 
	\label{eq6}
	{P}'(p)= \beta D\left ( p \right )+\left ( 1-\beta  \right ){C}'\left ( p \right )
\end{equation}	
Where $\beta$ is the weight coefficient, a real number within the range of 0 to 1.
\subsection{Selection of the Best Matching Block}
In the Criminisi algorithm, the window size of the sample block is fixed at 9 pixels $\times$  9 pixels. However, when the image contains rich detail information, a smaller window can make the repair more refined. Conversely, when there is less detail information, the use of a larger window can avoid block effects. To adapt to the variation in detail information in different images, this paper introduces the concept of local variance and dynamically adjusts the window size of the sample block. The local variance of the image can effectively reflect the richness of detail information in a local region. The larger the local variance, the richer the detail information in the local region. Specifically, by calculating the local variance of the surrounding area for each pixel, the algorithm can adjust the window size of the sample block based on the magnitude of the local variance. The adjustment of the window size of the sample block is expressed as Equation (7). 
\begin{equation} 
	\label{eq7}
	SZ(p)= \begin{cases} 9& \text meanS> Std(p) \\  5& \text meanS\leq Std(p) \end{cases}
\end{equation}	
Where $SZ(p)$ represents the window size of the sample block, Smean is the mean of the variances of known region pixels within all windows centered on the boundary pixels of the high-gloss area to be removed in the WCE image, and $Std(p)$ represents the variance of known region pixels within the window centered on point $p$.

After finding the pixel p with the highest priority among the boundary pixels, 
the algorithm then searches for the best matching block in the known region. 
The determination of the best matching block $\Psi _{\hat{p}}$  in the Criminisi algorithm is expressed as Equation (8).
\begin{equation} 
	\label{eq8}
	\Psi _{\hat{p}}= arg min_{\Psi q\in \Phi  } SSD(\Psi _{p},\Psi _{q})
\end{equation}	
Where $SSD(\Psi _{p},\Psi _{q})$ represents the squared differences in RGB color channels between two sample blocks, and its expression is given by Equation (9).
\begin{multline} 
	\label{eq9}
	SSD(\Psi _{p},\Psi _{q})= \\\sqrt{(R_{a}-R_{b})^{2}+(G_{a}-G_{b})^{2}+(B_{a}-B_{b})^{2}},  a\in \Psi _{p}, b\in \Psi _{q}
\end{multline}	

The texture gradient between images is continuous, and the closer the distance between sample blocks, the higher their similarity. To obtain a more accurate best-matching block, this paper has improved the selection of the best-matching block in the Criminisi algorithm in Equation (8), taking into account the distance between sample blocks when obtaining the best-matching block. The improved best-matching block is given by Equation (10).
\begin{equation} 
	\label{eq10}
	\Psi _{\hat{p}}^{{}'} =arg min_{\Psi q\in \Phi  }SSD(\Psi _{p},\Psi _{q})\cdot (p,q)
\end{equation}	
Where $L(p, q)$ represents the distance between points $p$ and $q$, and its expression is given by Equation (11).
\begin{equation} 
	\label{eq11}
	L(p,q)= \sqrt{(p(x)-q(x))^{2}+(p(x)-q(x))^{2}}
\end{equation}	
Where $x$ and $y$ denote the pixel indices of the image.
\subsection{Texture Information Inpainting}
Once the best-matching block is found, the pixel values of the center pixels in the best-matching block are copied to the specular highlight removal region of the highest-priority pixel p, as expressed in Equation (12). 
\begin{equation} 
	\label{eq12}
	\Psi _{p}= \Psi _{q},p\in (\Psi _{p}\cap \Omega  )
\end{equation}	
Where, $\Psi _{p} $ represents the pixel values of the highest-priority pixel $p$, 
i.e., the pixel values in the specular highlight removal region. $\Psi _{p}$ represents the center pixel value in the best-matching block.

Simultaneously, update the specular highlight region Mask matrix. The expression is shown in Equation (13).
\begin{equation} 
	\label{eq13}
	Mask(p)= 0,p\in (\Psi _{p}\cap \Omega  )
\end{equation}	

Repeat the above steps until the entire specular highlight region Mask matrix is all 0.

Finally, to prevent excessive differences at the boundary between the specular highlight region and the known region, this paper replaces the center pixel value in the 9-pixel $\times$ 9-pixel window centered on all pixels in the specular highlight region boundary with the average value of all pixels in the window. This completes the specular highlight removal from the WCE specular highlight image.
\section{Experimental Results and Analysis}
\subsection{Selection of Experimental Data}
To validate the effectiveness of the proposed method for specular highlight removal in WCE images, the CVC-ClinicSpec dataset \cite{ref26} was employed as the experimental dataset. This dataset includes capsule endoscopy images with annotated specular highlight regions by experienced medical professionals. Figure 3 illustrates examples from the CVC-ClinicSpec dataset. Due to the limited presence of small intestine highlight data in this dataset, twenty additional images with small intestine highlight regions were selected from a dataset provided by the company Huawei Tai Lai and incorporated into the experiment. The highlight regions in these supplementary images were manually labeled by trained professionals. Figure 4 provides examples of the supplemented dataset. 
\begin{figure}[H]	
	\centering
	\begin{subfigure}{0.35\linewidth}
		\centering
		\includegraphics[width=0.95\linewidth]{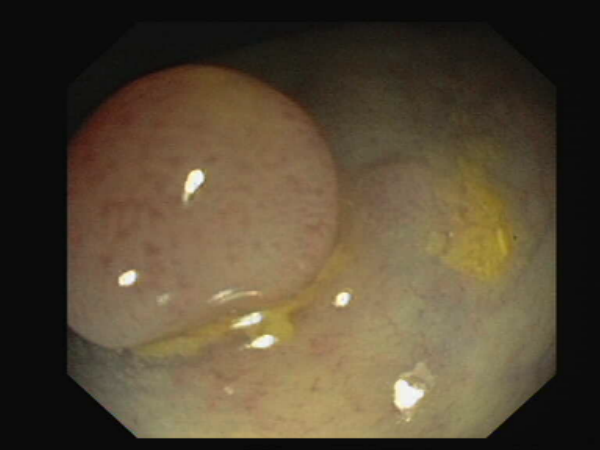}
		\caption{original}
		\label{Fig31}%
	\end{subfigure}
	\centering
	\begin{subfigure}{0.35\linewidth}
		\centering
		\includegraphics[width=0.95\linewidth]{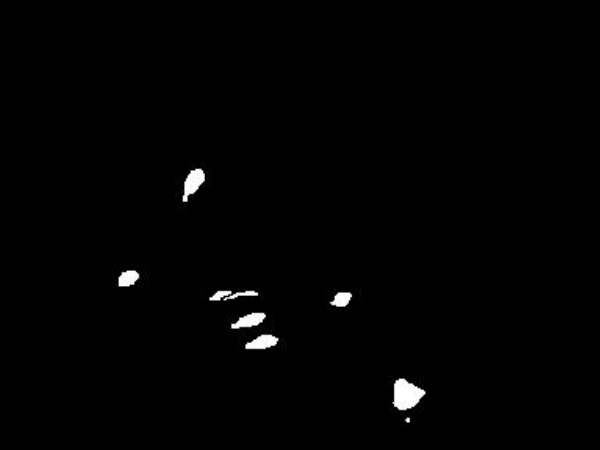}
		\caption{ highlight marker}
		\label{Fig32}
	\end{subfigure}
	\caption{Example of the CVC-ClinicSpec Database.}
\end{figure}

\begin{figure}[H]	
	\centering
	\begin{subfigure}{0.3\linewidth}
		\centering
		\includegraphics[width=0.95\linewidth]{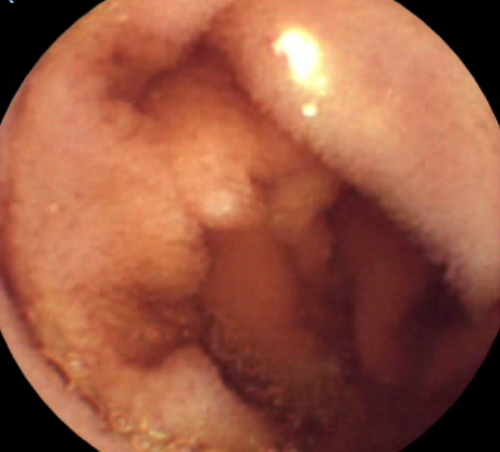}
		\caption{original}
		\label{Fig41}
	\end{subfigure}
	\centering
	\begin{subfigure}{0.3\linewidth}
		\centering
		\includegraphics[width=0.95\linewidth]{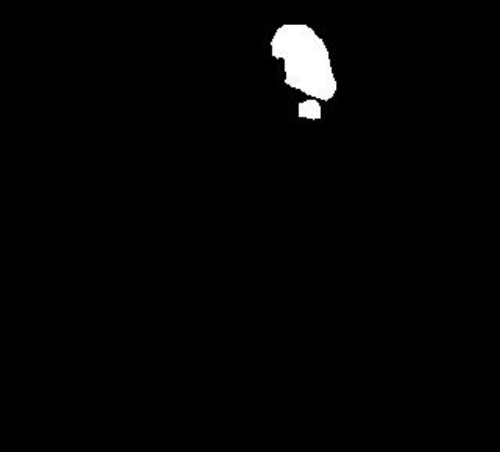}
		\caption{ highlight marker}
		\label{Fig42}
	\end{subfigure}
	\caption{Supplementary Database Example.}
\end{figure}
\subsection{Supplementary Database Example}
In this study, the standard deviation (Std), mean, and coefficient of variation (COV) of the specular highlight removal region in the WCE images were adopted as evaluation metrics for the proposed highlight removal method.

Std represents the degree to which pixels deviate from the mean in the entire image. In the context of highlight removal, the standard deviation can be used to measure the dispersion of the highlight regions in the image. A smaller standard deviation may indicate that the highlight region is more concentrated, while a larger standard deviation may suggest a more dispersed highlight region. The calculation is expressed as follows in Equation 14.

\begin{equation} 
	\label{eq14}
	Std= \sqrt{\frac{\sum \begin{matrix} M\\ i=1 \end{matrix}\sum \begin{matrix}N\\ j=1 \end{matrix}(I(x,y)-Mean)^{2}}{MN}}
\end{equation}	
Where $I(x,y)$ is the grayscale value at pixel $(x,y)$, and $M \times N$ represents the dimensions of the image. Mean is the average grayscale value of all pixels in the image.
Mean reflects the overall brightness of the image. For a grayscale image $I(x,y)$ with dimensions $M \times N$, the mean is calculated as follows in Equation 15.
\begin{equation} 
	\label{eq15}
	Mean= \frac{\sum \begin{matrix}M\\ i=1 \end{matrix}\sum \begin{matrix} N\\ j=1 \end{matrix}I(x,y)}{MN}
\end{equation}	

$COV$ represents the ratio of the standard deviation to the mean value in the region of the high-light-removed result image corresponding to the high-light area in the original image. It is used to measure the uniformity of intensity in a specific region. A smaller COV value indicates a better high-light removal effect. Its calculation is expressed as Equation (16).
\begin{equation} 
	\label{eq16}
	COV= \frac{Std}{Mean}
\end{equation}	
Where $Std$ is the standard deviation, and $Mean$ is the mean value. 

By utilizing these three metrics, a comprehensive assessment can be conducted to evaluate the impact of the high-light removal method on the image, thus providing a quantitative analysis of the effectiveness of the proposed approach. 
\subsection{Experimental Parameters}
In the process of improving the calculation of pixel priority, as shown in the above formula (6), 
where $\beta$  is the weight coefficient used to balance the relationship between the feature term and confidence. 
This coefficient has a range of real numbers between 0 and 1. The specific value can be determined through experiments 
to ensure the optimal balance between feature term and confidence in different scenarios. Figure 5 shows the results 
of WCE image highlight removal when $\beta$ takes different values. Figure 6 shows the evaluation metric values corresponding 
to the WCE image highlight removal results when $\beta$ takes different values. In Figure 6, it can be observed that
 when $\beta$ is set to 0.8, the Std and Cov values of the WCE image highlight removal area show a noticeable decrease, 
 indicating that the intensity in this region becomes more uniform. When $\beta$ is set to values greater than 0.8, 
 Std and Cov, on the contrary, increase, suggesting an increase in brightness differences in the image. 
 By observing the highlight removal results in Figure 5, it is evident that the removal effect is optimal 
 when $\beta$ is set to 0.8. Therefore, when $\beta$ is set to 0.8,
 it effectively balances the impact of the feature term and confidence, and this value is adopted in this paper.
 \begin{figure}[H]	
	\centering
	\begin{subfigure}{0.25\linewidth}
		\centering
		\includegraphics[width=0.95\linewidth]{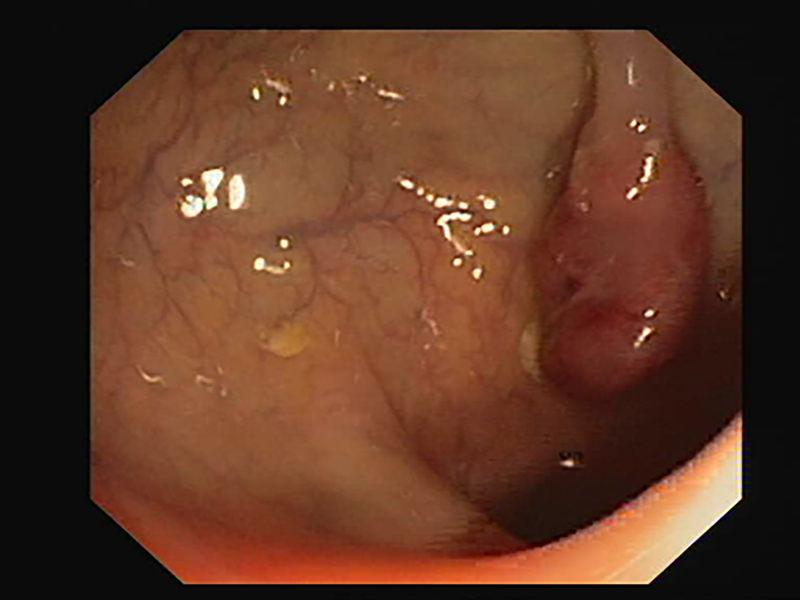}
		\caption{original}
		\label{Fig51}%
	\end{subfigure}
	\centering
	\begin{subfigure}{0.25\linewidth}
		\centering
		\includegraphics[width=0.95\linewidth]{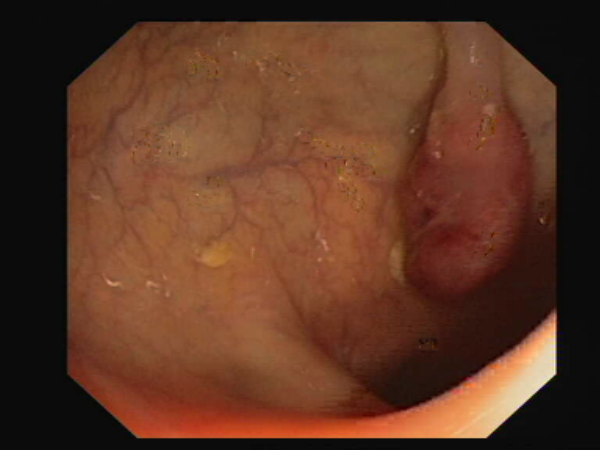}
		\caption{$\beta$=0.2}
		\label{Fig52}
	\end{subfigure}
	\begin{subfigure}{0.25\linewidth}
		\centering
		\includegraphics[width=0.95\linewidth]{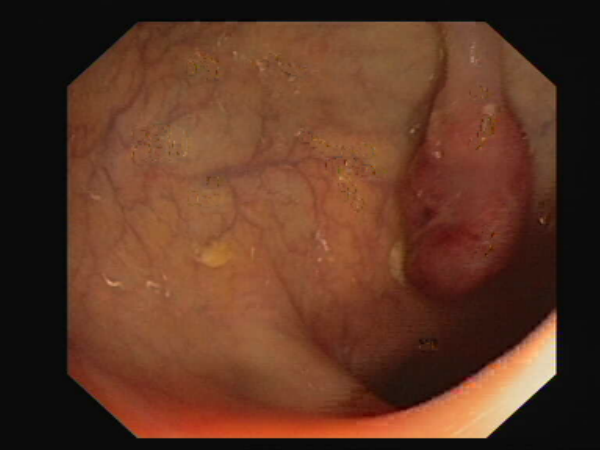}
		\caption{ $\beta$=0.4}
		\label{Fig53}
	\end{subfigure}
	\centering
	\begin{subfigure}{0.25\linewidth}
		\centering
		\includegraphics[width=0.95\linewidth]{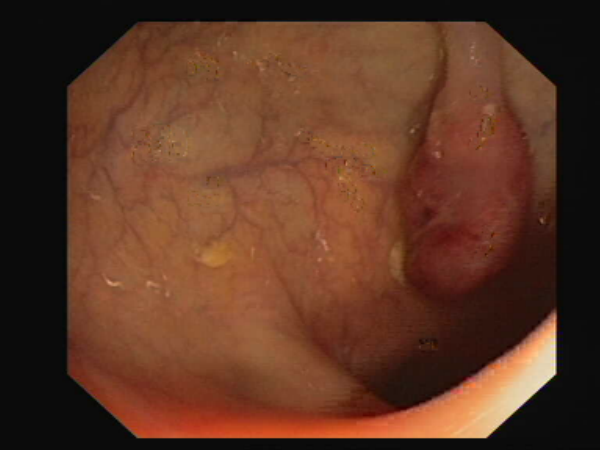}
		\caption{$\beta$=0.6}
		\label{Fig54}%
	\end{subfigure}
	\centering
	\begin{subfigure}{0.25\linewidth}
		\centering
		\includegraphics[width=0.95\linewidth]{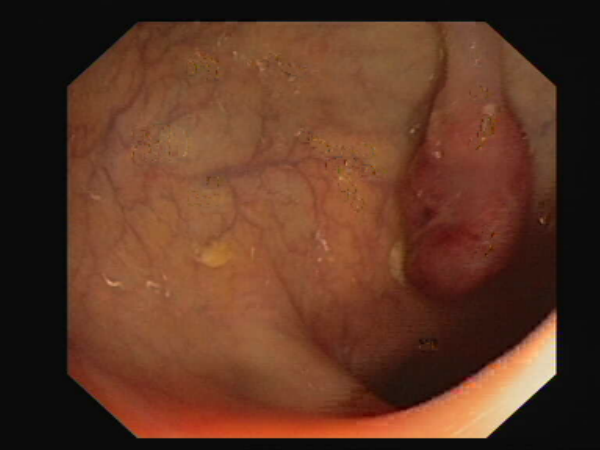}
		\caption{$\beta$=0.8}
		\label{Fig55}
	\end{subfigure}
	\begin{subfigure}{0.25\linewidth}
		\centering
		\includegraphics[width=0.95\linewidth]{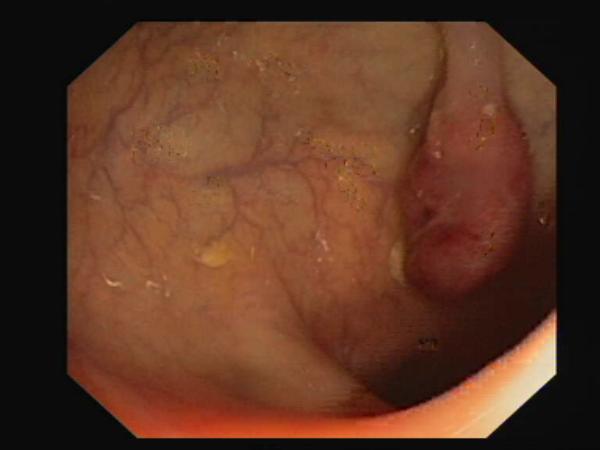}
		\caption{ $\beta$=0.9}
		\label{Fig56}
	\end{subfigure}
	\caption{Results of Highlight Removal in WCE Images with Different Values of Parameter $\beta$.}
\end{figure}

\begin{figure}[H]	
	\centering
	\begin{subfigure}{0.32\linewidth}
		\centering
		\includegraphics[width=0.99\linewidth]{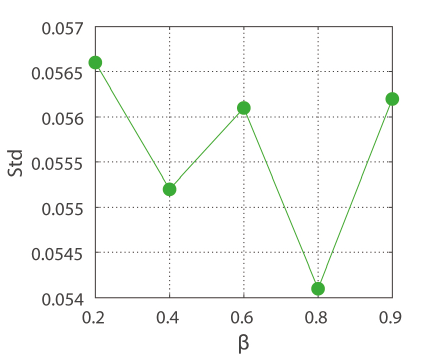}
		\caption{Std}
		\label{Fig61}%
	\end{subfigure}
	\centering
	\begin{subfigure}{0.32\linewidth}
		\centering
		\includegraphics[width=0.99\linewidth]{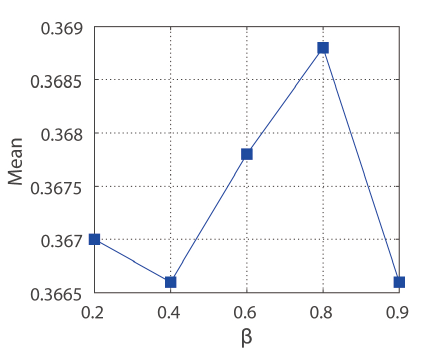}
		\caption{Mean}
		\label{Fig62}
	\end{subfigure}
	\begin{subfigure}{0.32\linewidth}
		\centering
		\includegraphics[width=0.99\linewidth]{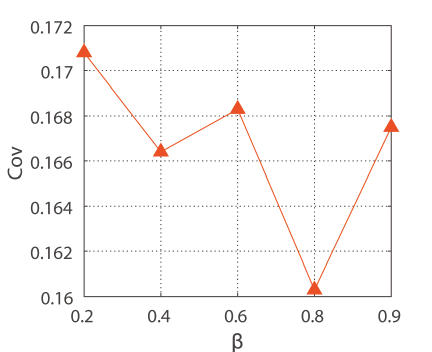}
		\caption{Cov}
		\label{Fig63}
	\end{subfigure}
	\caption{Example of the CVC-ClinicSpec Database.}
\end{figure}

 To verify the effectiveness of the improved pixel priority calculation method $(p1)$ and the improved best matching block selection method (p2) for highlight removal, ablation experiments were conducted. Figure 7 shows the experimental results on real digestive tract images, and Table 1 presents the corresponding evaluation metric values, with the minimum value of each metric bolded for clarity. In Table 1, it can be observed that the Cov values after improving pixel priority calculation and best matching block selection for images in Figure 7(a), (i) are lower than the results of highlight removal when only pixel priority calculation and best matching block selection are improved. Additionally, although the Cov value after simultaneously improving pixel priority calculation and best matching block selection for the image in Figure 7(e) is not optimal, in the green boxes of Figure 7(f), (g), it can be noticed that only when pixel priority and best matching block are improved, there are noticeable differences between the highlight removal area pixels and the surrounding area pixels. Therefore, simultaneous improvements in pixel priority calculation and best matching block selection result in more uniform intensity and more thorough highlight removal in the highlight removal area of WCE images.
 \begin{table}[!hptb]
	\newcommand{\tabincell}[2]{\begin{tabular}{@{}#1@{}}#2\end{tabular}}
	\centering
	\caption{Pixel priority calculation improvement and optimal matching block selection improvement ablation experiment evaluation indicator values}
	\setlength{\tabcolsep}{3pt}
	\renewcommand{\arraystretch}{1.2}
	\begin{tabular}{ccccc}  
	  \toprule
	  \multicolumn{2}{c}{Ablation portion} &
	 Std & Mean & Cov \\ \hline
	 \multirow{3}{*}{\tabincell{c}{Pixel priority calculation \\improvement  (p1)}}& image (a)  & 0.0115 & \textbf{0.0573} & 0.2013\\
	 & image (e)  & \textbf{0.0221} & 0.1009 & 0.2186 \\ 
	 & image (i)  & 0.0286 & 0.0747 & 0.3834 \\ \cline{2-5}
	 \multirow{3}{*}{\tabincell{c}{Optimal matching block \\selection  improvement (p2)}}& image (a) & 0.0108&0.0576&	0.1881\\
	 & image (e)&0.0254&0.1008&0.2523 \\ 
	 & image (i)&0.0304&0.0755&0.4027\\ \cline{2-5}
	 \multirow{3}{*}{p1+p2}& image (a) & \textbf{0.0096}&0.0588&\textbf{0.1642}\\
	 & image (e)&0.0223&\textbf{0.0952}&0.2338\\ 
	 & image (c)&\textbf{0.0274}&0.0729&\textbf{0.3762}\\ 
	  \bottomrule
	\end{tabular}
	
	\label{tab1}
  \end{table}   

  \begin{figure}[!hptb]
	\centering
	\begin{subfigure}{0.24\linewidth}
		\centering
		\includegraphics[width=0.95\linewidth]{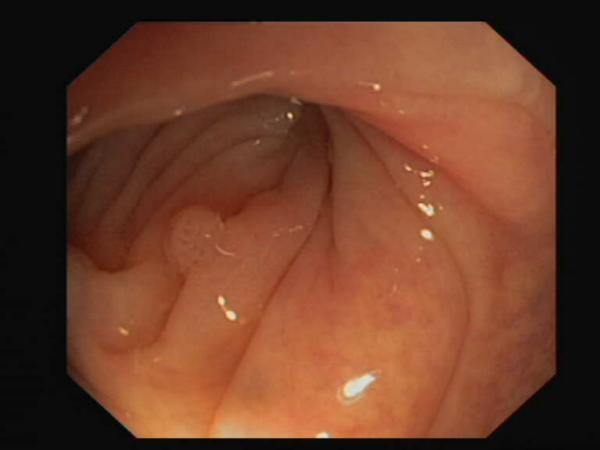}
		\caption{original}
		\label{Fig71}
	\end{subfigure}
	\centering
	\begin{subfigure}{0.24\linewidth}
		\centering
		\includegraphics[width=0.95\linewidth]{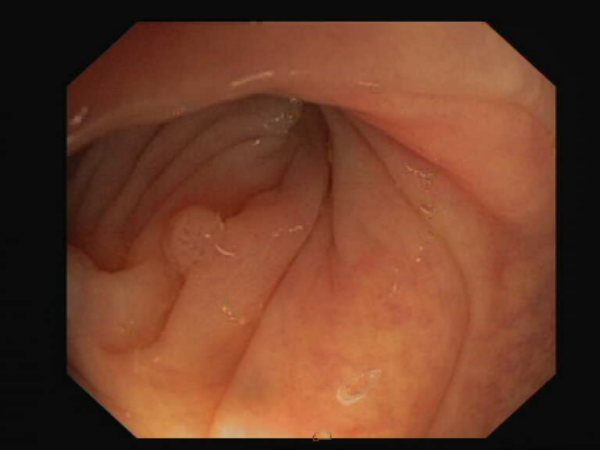}
		\caption{p1}
		\label{Fig72}
	\end{subfigure}
	\centering
	\begin{subfigure}{0.24\linewidth}
		\centering
		\includegraphics[width=0.95\linewidth]{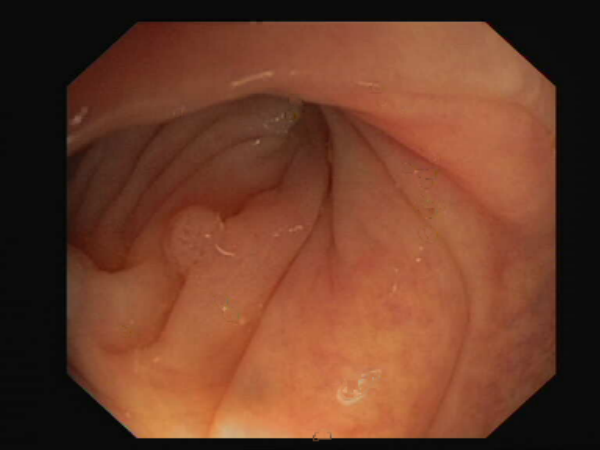}
		\caption{p2}
		\label{Fig73}
	\end{subfigure}
	\centering
	\begin{subfigure}{0.24\linewidth}
		\centering
		\includegraphics[width=0.95\linewidth]{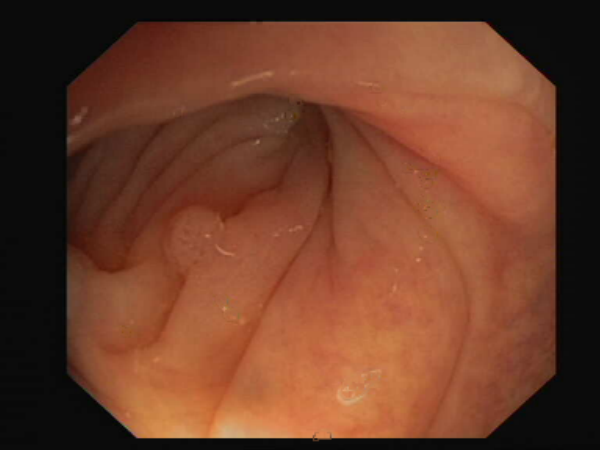}
		\caption{p1+p2}
		\label{Fig74}
	\end{subfigure}
	\centering
	\begin{subfigure}{0.24\linewidth}
		\centering
		\includegraphics[width=0.95\linewidth]{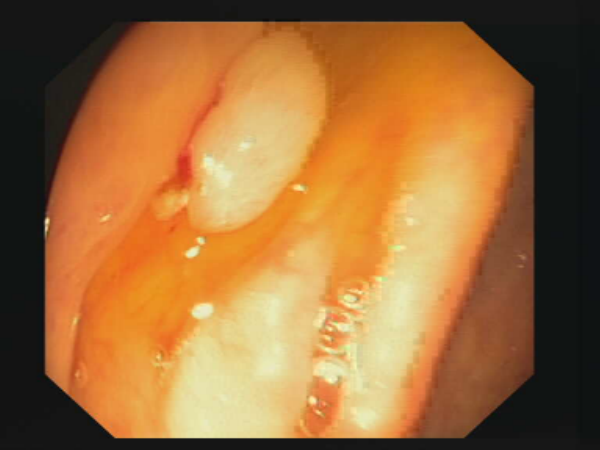}
		\caption{original}
		\label{Fig125}
	\end{subfigure}
	\centering
	\begin{subfigure}{0.24\linewidth}
		\centering
		\includegraphics[width=0.95\linewidth]{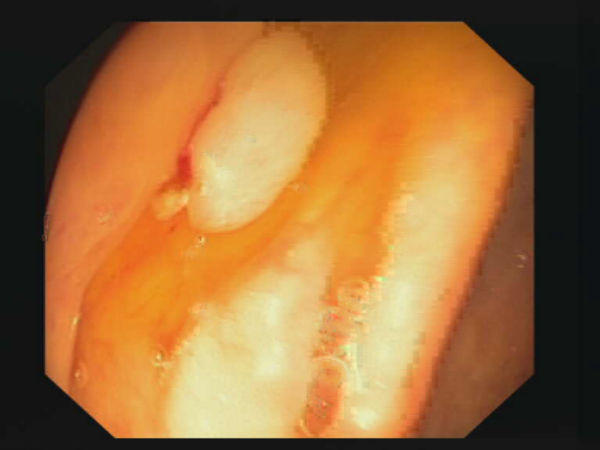}
		\caption{pl}
		\label{Fig76}
	\end{subfigure}
	\centering
	\begin{subfigure}{0.24\linewidth}
		\centering
		\includegraphics[width=0.95\linewidth]{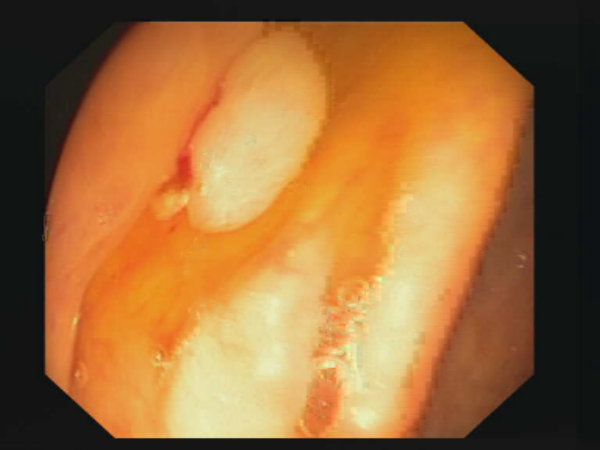}
		\caption{p2}
		\label{Fig77}%
	\end{subfigure}
	\centering
	\begin{subfigure}{0.24\linewidth}
		\centering
		\includegraphics[width=0.95\linewidth]{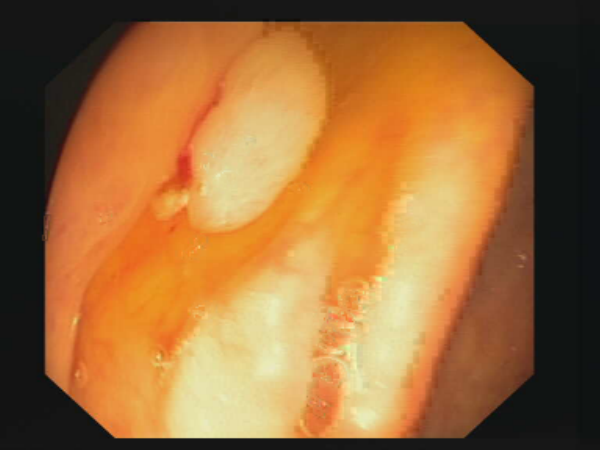}
		\caption{p1+p2}
		\label{Fig78}
	\end{subfigure}
	\centering
	\begin{subfigure}{0.24\linewidth}
		\centering
		\includegraphics[width=0.95\linewidth]{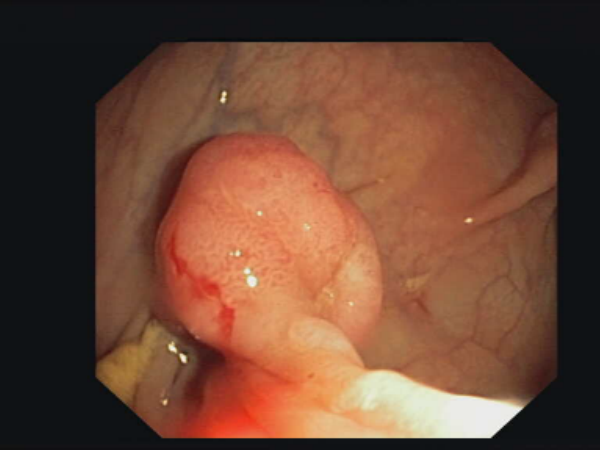}
		\caption{original}
		\label{Fig79}
	\end{subfigure}
	\centering
	\begin{subfigure}{0.24\linewidth}
		\centering
		\includegraphics[width=0.95\linewidth]{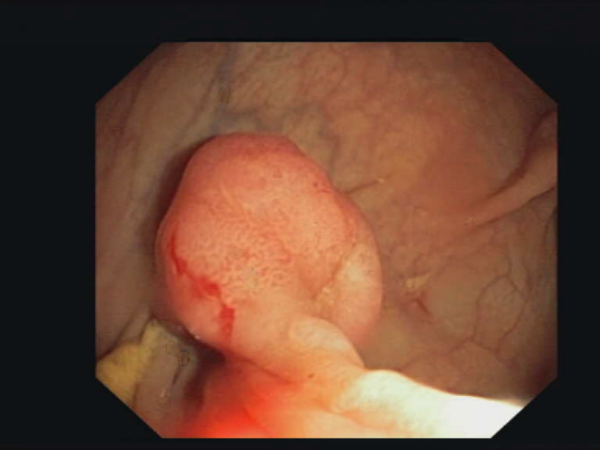}
		\caption{p1}
		\label{Fig710}
	\end{subfigure}
	\centering
	\begin{subfigure}{0.24\linewidth}
		\centering
		\includegraphics[width=0.95\linewidth]{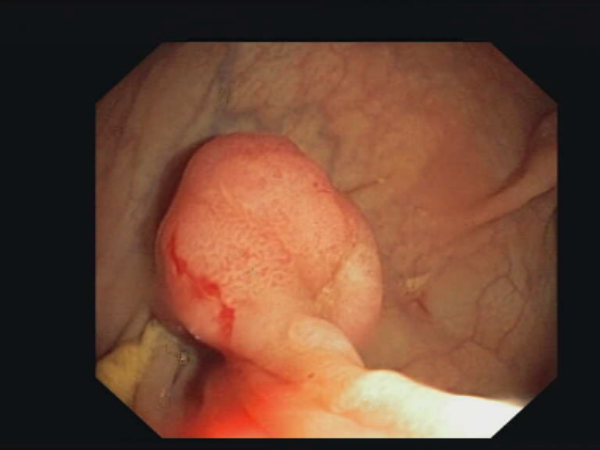}
		\caption{p2}
		\label{Fig711}
	\end{subfigure}
	\centering
	\begin{subfigure}{0.24\linewidth}
		\centering
		\includegraphics[width=0.95\linewidth]{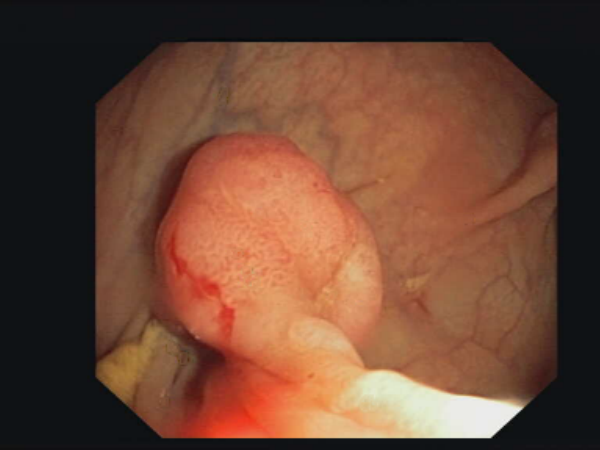}
		\caption{p1+p2}
		\label{Fig712}
	\end{subfigure}
	\caption{ Ablation Experiment of Pixel Priority Calculation Improvement and Optimal Matching Block Selection Improvement.}
	\label{Fig7}
\end{figure}
  \subsection{Experimental Comparative Analysis}
  To evaluate the effectiveness of our method, we compared it with DeepGin [22] and the Criminisi algorithm [13]. Figure 8 shows the highlight removal results of DeepGin, the Criminisi algorithm, and our method for WCE images. Figure 8(a) represents a human small intestine image, while Figures 8(e), (i), (m), and (q) represent human colon and stomach images. Table 2 provides the evaluation metric values for the processed results of Figures 8(a), (e), (i), (m), and (q).

  In Table 2, for Figure 8(a), the Std and Mean values of the image processed by DeepGin in Figure 8(b) are higher than our method, indicating larger pixel value fluctuations and higher brightness in the removed highlight region. It can be observed in the green box in Figure 8(b) that the highlight removal result of DeepGin still has overall higher brightness than its surrounding areas. For the image processed by the Criminisi algorithm in Figure 8(c), the Std value is higher than our method, and the Mean value is lower, suggesting significant pixel value fluctuations and lower brightness in the removed highlight region. In the green box in Figure 8(c), it is noticeable that the pixels in the highlight removal region after Crinimisi algorithm processing differ significantly from the surrounding pixels. In Table 2, the Std and Cov values for the image processed by our method in Figure 8(d) are both lower than the other two methods, indicating smaller pixel value fluctuations and better highlight removal effects. Additionally, in Figure 8(d), it can be observed that the highlight removal region after our method is similar to nearby pixels and exhibits continuous texture.

  In Table 2, for the highlight removal results of Figures 8(e), (i), (m), our method in Figures 8(h), (l), (p) has the lowest Std and Cov values, indicating smaller pixel value fluctuations and better highlight removal effects. In the green boxes of Figures 8(h), (l), (p), it can be observed that the highlight removal areas repaired by our method are similar to nearby pixels and have coherent textures. For the image processed by DeepGin in Figure 8(j), the Std value is higher than our method, and the Mean value is lower, suggesting larger pixel value fluctuations and lower brightness in the removed highlight region. Moreover, for the image processed by Crinimisi algorithm in Figure 8(k), both the Std and Mean values are higher than our method. In the green boxes of Figures 8(j), (k), it is noticeable that the highlight removal regions after DeepGin and Crinimisi algorithm processing have higher brightness than their surroundings and exhibit texture artifacts. In Figure 8(l), it can be observed that the highlight removal region repaired by our method is similar to nearby pixels and has coherent textures.

  In Table 2, for the image processed by DeepGin in Figure 8(r),
   both the Std and Mean values are higher than our method, 
   indicating larger pixel value fluctuations and higher brightness in the removed highlight region. In Figure 8(r), 
   it can be observed that the highlight removal is incomplete in the green box. 
   For the image processed by the Criminisi algorithm and our method in Figure 8(q), the Std value is lower than our method,
    and the Mean value is higher than our method, suggesting smaller pixel value fluctuations and higher brightness 
	in the removed highlight region. In the blue box of Figure 8(s), it is noticeable that the highlight removal region 
	after Crinimisi algorithm processing differs significantly from the surrounding pixels. In Figure 8(t), 
	it can be observed that the highlight removal region repaired by our method is similar to nearby pixels and has a smooth texture.

  \begin{figure}[!hptb]

	\begin{subfigure}{0.24\linewidth}
		\centering
		\includegraphics[width=0.95\linewidth]{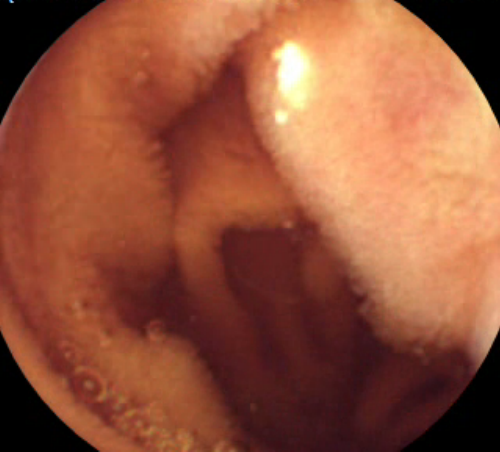}
		\caption{original}
		\label{Fig81}
	\end{subfigure}
	\centering
	\begin{subfigure}{0.24\linewidth}
		\centering
		\includegraphics[width=0.95\linewidth]{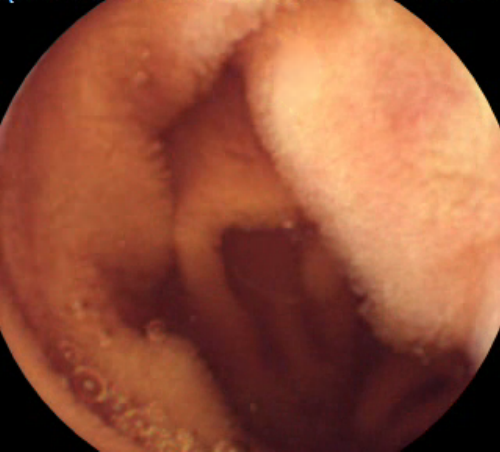}
		\caption{DeepGin}
		\label{Fig82}
	\end{subfigure}
	\centering
	\begin{subfigure}{0.24\linewidth}
		\centering
		\includegraphics[width=0.95\linewidth]{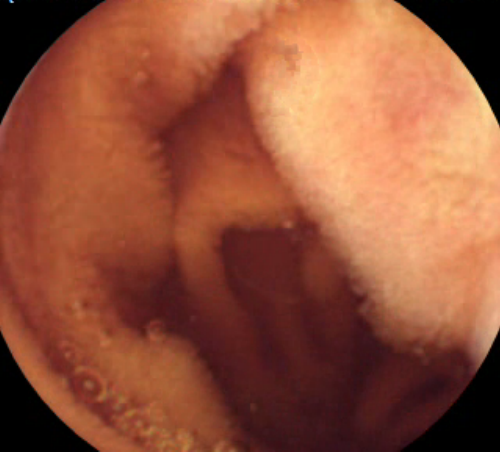}
		\caption{Criminisi}
		\label{Fig83}
	\end{subfigure}
	\centering
	\begin{subfigure}{0.24\linewidth}
		\centering
		\includegraphics[width=0.95\linewidth]{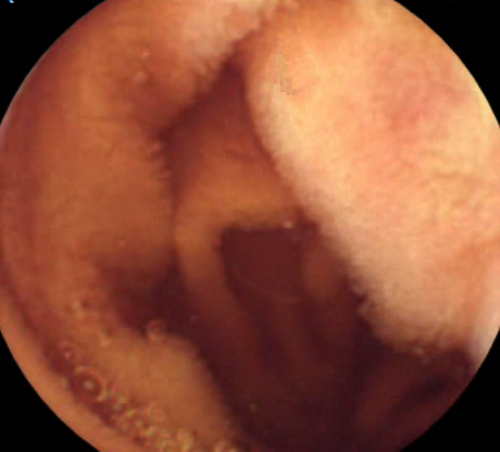}
		\caption{proposed}
		\label{Fig84}
	\end{subfigure}
	\centering
	\begin{subfigure}{0.24\linewidth}
		\centering
		\includegraphics[width=0.95\linewidth]{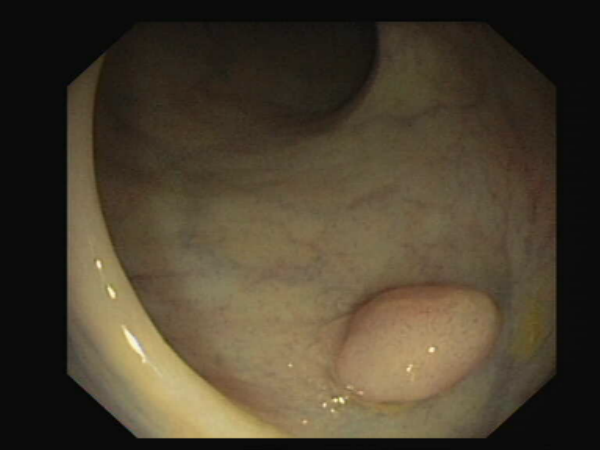}
		\caption{original}
		\label{Fig85}
	\end{subfigure}
	\centering
	\begin{subfigure}{0.24\linewidth}
		\centering
		\includegraphics[width=0.95\linewidth]{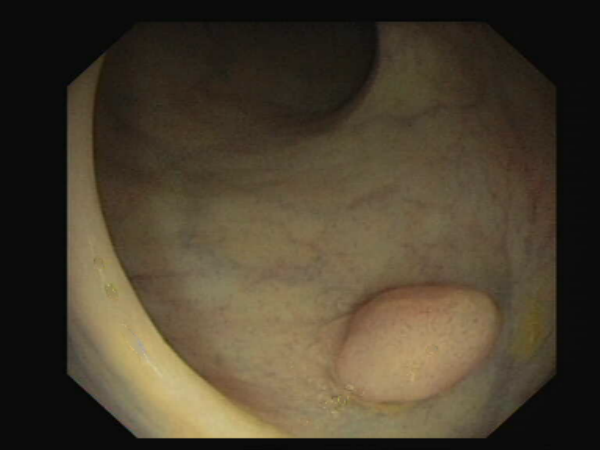}
		\caption{DeepGin}
		\label{Fig86}
	\end{subfigure}
	\centering
	\begin{subfigure}{0.24\linewidth}
		\centering
		\includegraphics[width=0.95\linewidth]{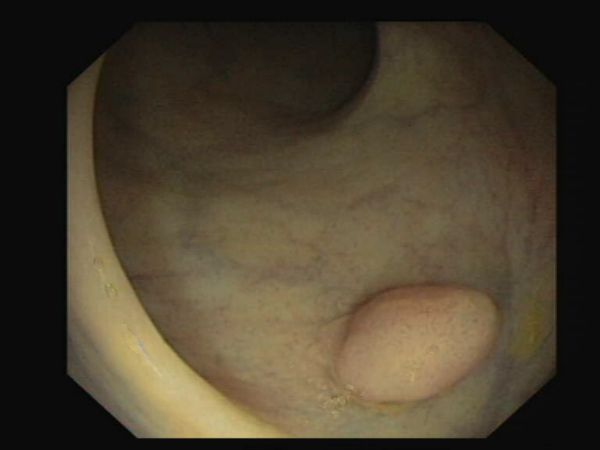}
		\caption{Criminisi}
		\label{Fig87}
	\end{subfigure}
	\centering
	\begin{subfigure}{0.24\linewidth}
		\centering
		\includegraphics[width=0.95\linewidth]{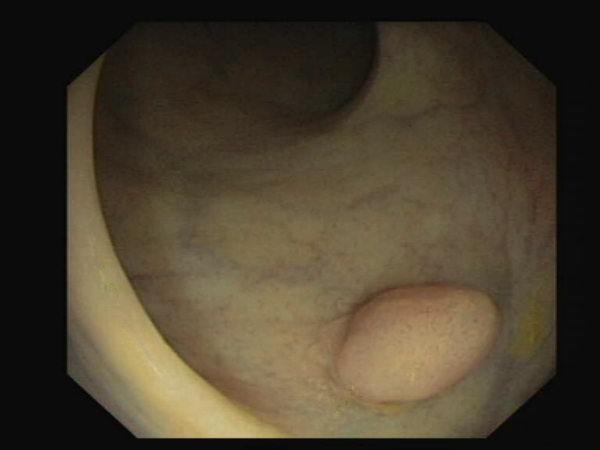}
		\caption{proposed}
		\label{Fig88}
	\end{subfigure}
	\centering
	\begin{subfigure}{0.24\linewidth}
		\centering
		\includegraphics[width=0.95\linewidth]{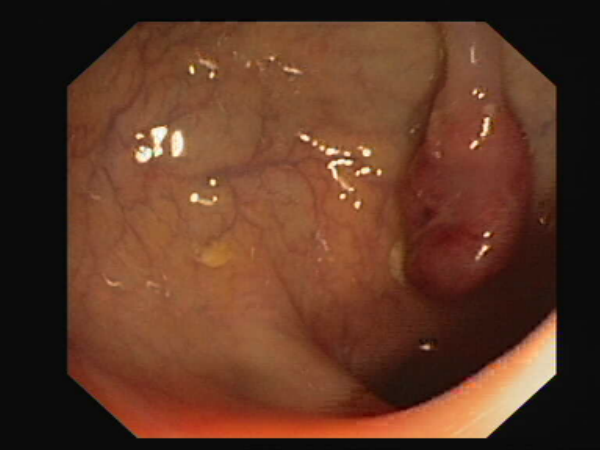}
		\caption{original}
		\label{Fig89}
	\end{subfigure}
	\centering
	\begin{subfigure}{0.24\linewidth}
		\centering
		\includegraphics[width=0.95\linewidth]{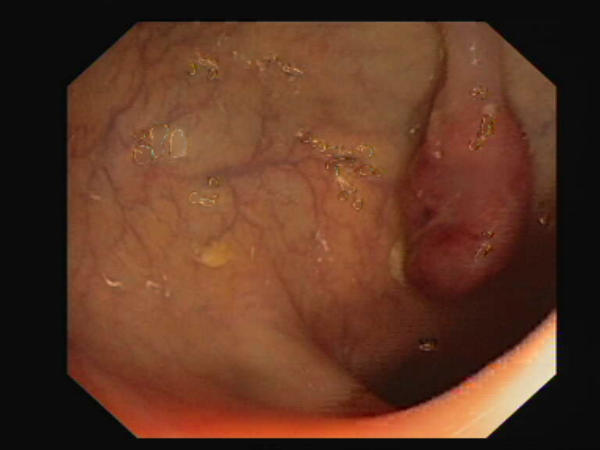}
		\caption{DeepGin}
		\label{Fig810}
	\end{subfigure}
	\centering
	\begin{subfigure}{0.24\linewidth}
		\centering
		\includegraphics[width=0.95\linewidth]{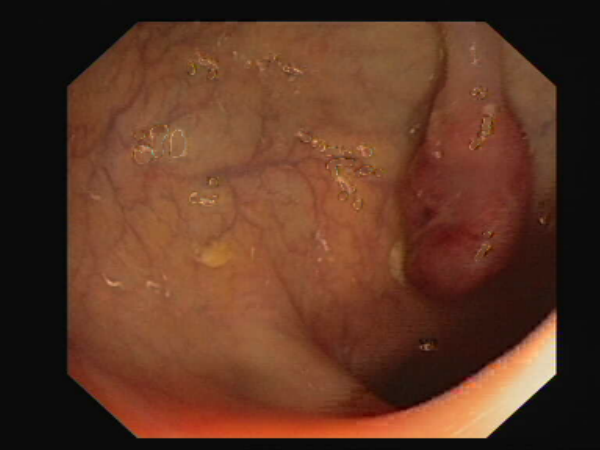}
		\caption{Criminisi}
		\label{Fig811}
	\end{subfigure}
	\centering
	\begin{subfigure}{0.24\linewidth}
		\centering
		\includegraphics[width=0.95\linewidth]{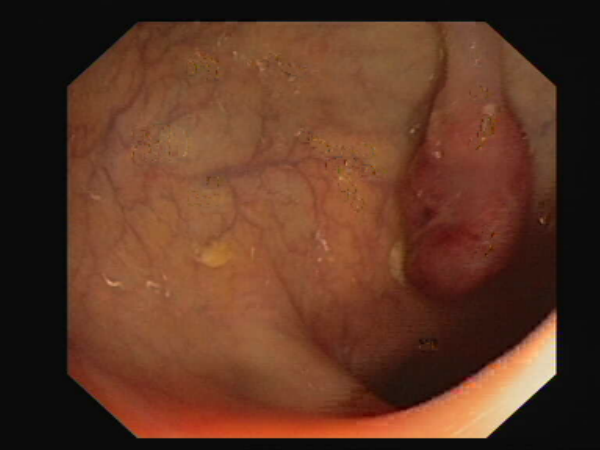}
		\caption{proposed}
		\label{Fig812}
	\end{subfigure}
	\centering
	\begin{subfigure}{0.24\linewidth}
		\centering
		\includegraphics[width=0.95\linewidth]{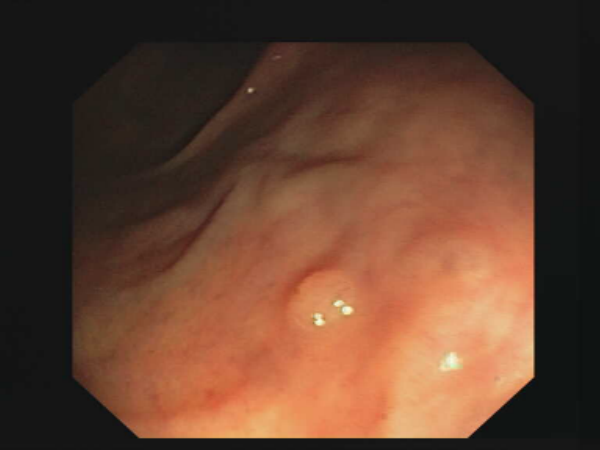}
		\caption{original}
		\label{Fig813}
	\end{subfigure}
	\centering
	\begin{subfigure}{0.24\linewidth}
		\centering
		\includegraphics[width=0.95\linewidth]{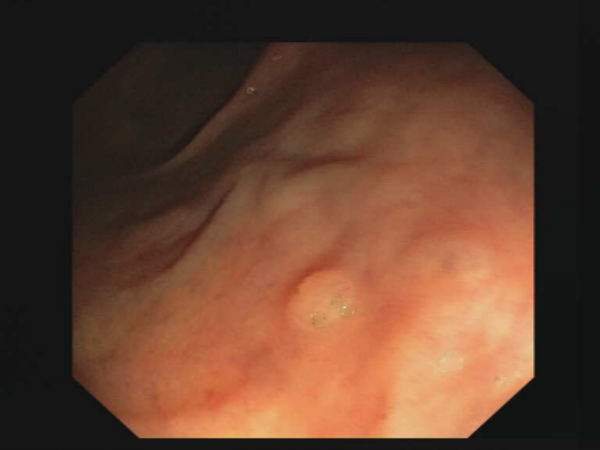}
		\caption{DeepGin}
		\label{Fig814}
	\end{subfigure}
	\centering
	\begin{subfigure}{0.24\linewidth}
		\centering
		\includegraphics[width=0.95\linewidth]{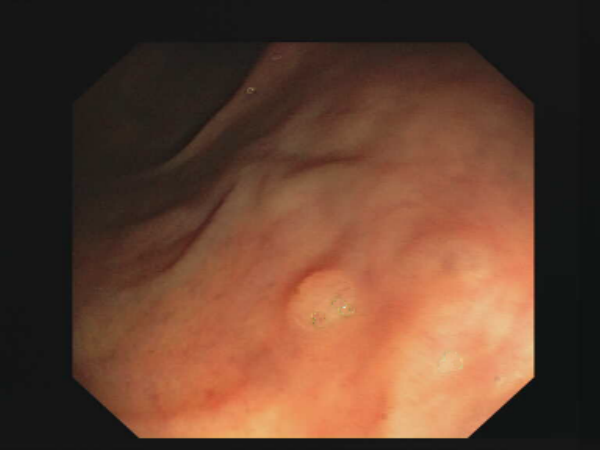}
		\caption{Criminisi}
		\label{Fig815}
	\end{subfigure}
	\centering
	\begin{subfigure}{0.24\linewidth}
		\centering
		\includegraphics[width=0.95\linewidth]{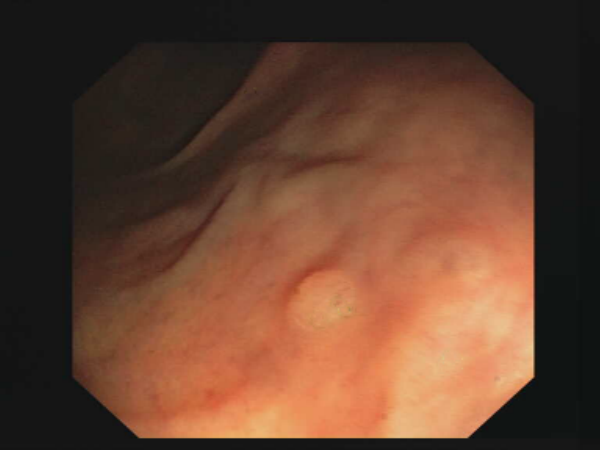}
		\caption{proposed}
		\label{Fig816}
	\end{subfigure}
	\caption{ Results of Highlight Removal using Different Methods, including DeepGin, Criminisi, and Our Own Method.}
	\label{Fig8}
\end{figure}
  \begin{table}[!hptb]
	\newcommand{\tabincell}[2]{\begin{tabular}{@{}#1@{}}#2\end{tabular}}
	\centering
	\caption{Select five different WCE images and evaluate the indicator values after highlight removal using different methods.}
	\setlength{\tabcolsep}{3pt}
	\renewcommand{\arraystretch}{1.2}
	\setlength{\tabcolsep}{10pt}
	\begin{tabular}{ccccc}  
	  \toprule
	 Images & Method &
	 Std & Mean & Cov \\ \hline
	 \multirow{3}{*}{\tabincell{c}{(a)}}& DeepGin  & 0.0597 & 0.7302 & 0.0918\\
	 & Criminisi  & 0.0742 & \textbf{0.6913} & 0.1191 \\ 
	 & proposed  & \textbf{0.0509} & 0.6939 & \textbf{0.0814} \\ \cline{2-5}
	 \multirow{3}{*}{\tabincell{c}{(e)}}& DeepGin & 0.0660&\textbf{0.5474}&0.1241\\
	 &Criminisi&0.0677&0.5653&0.1188 \\ 
	 & proposed&\textbf{0.0657}&0.5588&\textbf{0.1155}\\ \cline{2-5}
	 \multirow{3}{*}{\tabincell{c}{(i)}}& DeepGin & 0.0603&\textbf{0.3645}&0.1850\\
	 &Criminisi&0.0569&0.3830&0.1613 \\ 
	 & proposed&\textbf{0.0541}&0.3688&\textbf{0.1630}\\ \cline{2-5}
	 \multirow{3}{*}{\tabincell{c}{(m)}}& DeepGin & 0.1094&0.6359&0.1820\\
	 &Criminisi&0.1121&0.6418&0.1844 \\ 
	 & proposed&\textbf{0.1019}&\textbf{0.6333}&\textbf{0.1677}\\ \cline{2-5}
	 \multirow{3}{*}{\tabincell{c}{(q)}}& DeepGin & 0.25330&0.6351&0.4057\\
	 & Criminisi&\textbf{0.2358}&0.6283&\textbf{0.3872}\\ 
	 & proposed&0.2419&\textbf{0.6171}&0.4107\\ 
	  \bottomrule
	\end{tabular}
	
	\label{tab2}
  \end{table}   
	
Figure 9 shows the evaluation metric values for 60 WCE highlight images processed by DeepGin, the Criminisi algorithm, and our method. Table 3 presents the average evaluation metric values for these processed WCE highlight images. In Table 3, the average Std value after the removal of WCE highlight by our method is the smallest, indicating minimal pixel value fluctuations in the removed highlight region. Additionally, the average Cov value for our method is the lowest, indicating uniform intensity in the repaired highlight region and better overall restoration effects. These experiments collectively demonstrate that the pixel values of the highlight removal region after our method's processing are similar to nearby pixels, exhibiting continuous textures. The visual results after highlight removal are good, indicating that our method can effectively remove highlights from WCE highlight images.

\begin{figure}[H]	
	\centering
	\begin{subfigure}{0.45\linewidth}
		\centering
		\includegraphics[width=0.95\linewidth]{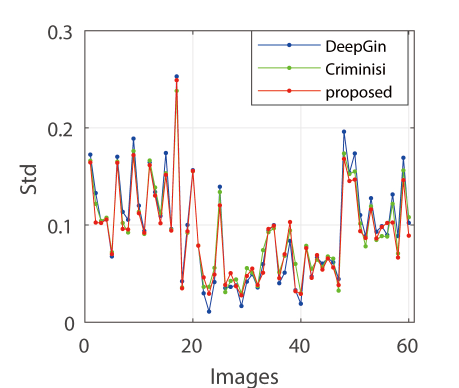}
		\caption{Std}
		\label{Fig91}%
	\end{subfigure}
	\centering
	\begin{subfigure}{0.45\linewidth}
		\centering
		\includegraphics[width=0.95\linewidth]{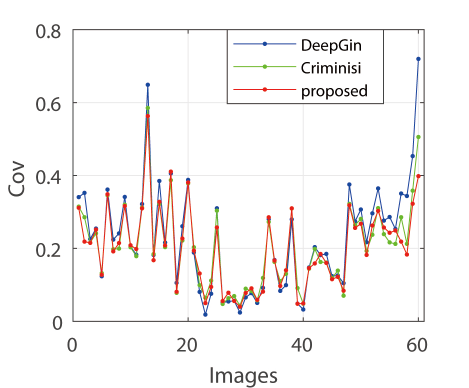}
		\caption{Mean}
		\label{Fig92}
	\end{subfigure}
	\begin{subfigure}{0.45\linewidth}
		\centering
		\includegraphics[width=0.95\linewidth]{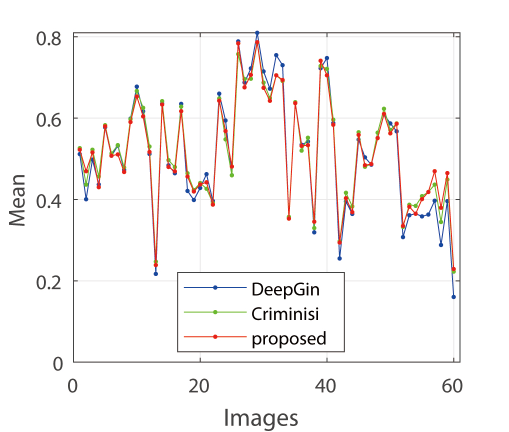}
		\caption{Cov}
		\label{Fig93}
	\end{subfigure}
	\caption{Select 60 different WCE images, evaluation metrics for individual images.}
\end{figure}

\begin{table}[!hptb]
	\centering
	\caption{Average Value of WCE Highlight Image Evaluation Indicators}
	\setlength{\tabcolsep}{15pt}
	\begin{tabular} {l *{3}{S[table-format=2.4]}}  
	  \toprule
	  Method & {Std} & {Mean} & {Cov} \\
	  \midrule
	  DeepGin & 0.0948 & \textbf{0.5132} & 0.2249 \\
	  Criminisi & 0.0937 & 0.5206 & 0.2070\\
	  proposed & \textbf{0.0908} & 0.5182 & \textbf{ 0.2013} \\
	
	  \bottomrule
	\end{tabular}
	
	\label{tab3}
  \end{table}    
\section{Conclusion}
Given that existing highlight removal methods are inadequate for the requirements in the human body cavity environment, this paper proposes a highlight removal method tailored for capsule endoscopy images. The method improves the confidence of the Criminisi algorithm by leveraging the characteristics of wireless capsule endoscopy (WCE) images, providing it with more known information. Furthermore, it adjusts the sample block window size based on the local variance of the highlight region edges and incorporates pixel distance factors in the computation of the best matching block to achieve more accurate results. Experimental results demonstrate that our method effectively removes specular highlights from WCE images while ensuring color similarity and texture continuity in the highlight removal region and its surroundings. Future work will focus on optimizing the WCE highlight removal method based on the results obtained in this paper, with a particular emphasis on minimizing runtime.





\end{document}